# Physics-Informed Neural Network based Damage Identification for Truss Railroad Bridges

Althaf Shajihan[1] | Kirill Mechitov[2] | Girish Chowdhary[3] | Billie F. Spencer Jr.[1]

[1] Department of Civil and Environmental Engineering, University of Illinois at Urbana-Champaign, Urbana, IL, USA

[2] StructureIQ, Champaign, IL, USA

[3] Coordinated Science Laboratory, University of Illinois at Urbana-Champaign, Urbana, IL, USA

Correspondence: Billie F. Spencer Jr., email: bfs@illinois.edu

## Abstract

Railroad bridges are a crucial component of the U.S. freight rail system, which moves over 40 percent of the nation's freight and plays a critical role in the economy. However, aging bridge infrastructure and increasing train traffic pose significant safety hazards and risk service disruptions. The U.S. rail network includes over 100,000 railroad bridges, averaging one every 1.4 miles of track, with steel bridges comprising over 50% of the network's total bridge length. Early identification and assessment of damage in these bridges remain challenging tasks. This study proposes a physics-informed neural network (PINN) based approach for damage identification in steel truss railroad bridges. The proposed approach employs an unsupervised learning approach, eliminating the need for large datasets typically required by supervised methods. The approach utilizes train wheel load data and bridge response during train crossing events as inputs for damage identification. The PINN model explicitly incorporates the governing differential equations of the linear time-varying (LTV) bridge-train system. Herein, this model employs a recurrent neural network (RNN) based architecture incorporating a custom Runge-Kutta (RK) integrator cell, designed for gradient-based learning. The proposed approach updates the bridge finite element model while also quantifying damage severity and localizing the affected structural members. A case study on the Calumet Bridge in Chicago, Illinois, with simulated damage scenarios, is used to demonstrate the model's effectiveness in identifying damage while maintaining low false-positive rates. Furthermore, the damage identification pipeline is designed to seamlessly integrate prior knowledge from inspections and drone surveys, also enabling context-aware updating and assessment of bridge's condition.

## 1. INTRODUCTION

Railroad bridges play a crucial role in the U.S. freight transportation network, and ensuring the structural integrity of these bridges is paramount for maintaining safe and efficient rail operations. The growing demands of rail transport, characterized by increased axle loads that often exceed the original design specifications for these bridges, have intensified the stress and fatigue on bridge structures, leading to accelerated deterioration. The composition of the U.S. railroad bridge inventory further underscores the critical importance of assessing the current state of bridges and developing effective damage identification strategies. The freight rail network is comprised of over

100,000 railroad bridges, more than half of which were constructed nearly a century ago [1]; more than 50% of the inventory, based on total bridge length in miles, consists of steel bridges [2]. Railroad bridge maintenance decisions are guided by insights gathered from bridge inspections, which include observations of bridge performance under revenue-service traffic conditions. Management of these aging structures presents unique challenges due to their susceptibility to fatigue, corrosion, and other time-dependent degradation mechanisms. A comprehensive Federal Railroad Administration (FRA) report [3] reveals that approximately 17% of reported railroad bridge failures from 1999 to 2010 were attributed to failed structural members. This alarming statistic not only highlights the inadequacies in current inspection approaches but also emphasizes the potential catastrophic consequences of undetected structural issues. Moreover, the economic ramifications of these failures are substantial, with an estimated average loss of $1.5 million per incident [3].

Routine condition assessments are therefore critical to identify deficient structures and prevent catastrophic failures. While visual inspection remains the most commonly used method for assessing railroad bridges, this approach has several inherent limitations [4,5]. Inspections are typically performed intermittently, increasing the risk of missing critical structural issues that can develop between intervals, especially in hard-to-reach areas of the bridge. These assessments are also highly subjective, relying on the inspector's experience and judgment, and are prone to human error. They are also time-consuming and costly, making frequent assessments impractical, which can result in subtle defects or deterioration being overlooked until they pose a risk to safety and performance. To address these issues, several researchers have investigated the use of structural health monitoring (SHM) strategies for railroad bridges. Critical railroad bridges of interest can be monitored using sensors to collect long-term and in-service data for the bridge. The sensors used in traditional methods typically consist of accelerometers, strain-gauges, and linear variable differential transformers (LVDTs) [5]. Ahmadi and Daneshjoo [6] implemented a monitoring system using accelerometers on a railroad bridge with known input train loads to assess key structural parameters. Banerji and Chikermane [7] collected strain and deflection measurements for in-service train loads on a masonry arch railroad bridge for model updating using a multi-response multi-parameter model updating approach. The emergence of deployable and low-cost wireless smart sensors (WSSs) for SHM, such as the Xnode [8] and Xnode-WSVS [9], has enabled multimodal and vision-based data acquisition in a scalable manner. Kim et al. [10] and Moreu et al. [11] used wireless strain gages mounted on the rail to estimate real-time train inputs loads and measured the associated responses with wireless accelerometers and magnetic strain gages installed on the railroad bridge. Hoang et al. [12] instrumented nine timber trestle railroad bridges using WSSs integrated with 4G-LTE support for remote data retrieval offering an autonomous end-to-end monitoring pipeline for railroad bridges. Yoon et al. developed an approach to measure deflection of in-service railroad bridges using a drone [13]. Furthermore, drone-based visual surveys have been investigated by researchers to also identify potential damage hot spots in railroad bridges, providing valuable information to complement sensor-based monitoring [14,15]. However, transitioning from data acquisition to effective condition assessment on railroad bridges remains a significant challenge.

Researchers have investigated various data-driven approaches for assessing railroad bridges. For example, Marasco et al. [16] applied sensor clustering and ARX modeling for damage localization and estimation using simulated acceleration data. Anastasia et al. [17] proposed an auto-regressive time-series model to extract damage-sensitive features from strain data on railroad bridges and performed damage identification through clustering. Other studies have employed optimization techniques, such as the Particle Swarm Optimization (PSO) algorithm [18], combined with statistical indicators for damage identification [19]. Hoang et al. [12] proposed qualitatively assessing timber-trestle railroad bridges using Gaussian Process Regression to assign safety ratings to in-service bridges. Zhan et al. [20] proposed a response sensitivity-based FE model updating approach for damage identification in numerical railroad bridge models. Furthermore, researchers have also investigated machine learning approaches for damage identification in railroad bridges. Hajializadeh [21] used simulated data from a drive-by train with a 2D model of a railroad bridge to perform damage detection using wavelet transform (WT) features and a CNN architecture. Shu et al. [22] presented an ANN-based supervised learning approach using statistical properties of simulated responses to identify damage in railroad bridge members. Lee et al. [23] employed a semi-supervised approach with cosine distance for anomaly detection, using wavelet-transformed acceleration data as image input for a CNN architecture to detect damage. Rageh et al. [24] proposed a supervised ANN model for fatigue damage prediction in steel railroad bridges. While these studies have contributed to the advancement in damage identification of railroad bridges, they do not directly provide an updated numerical model capable of predicting responses under unknown loading conditions. Additionally, the reliance on supervised deep learning models introduces two major challenges: the requirement for extensive training datasets and the lack of reliable performance when models are applied beyond their training data, particularly due to their 'black-box' nature. Thus, data-driven damage identification and model updating in railroad bridges have yet to mature.

Additionally, several challenges specific to railroad bridges increase the difficulty of performing model updating and damage identification, including: (a) Complexity of loading conditions: Railroad bridges are subjected to heavy, dynamic loads from trains, which can be comparable to the mass of the bridge itself. The moving train loads render the system dynamics as a linear time-varying (LTV) problem [10], making model updating more challenging than the commonly encountered linear time-invariant (LTI) problems. The interaction between the train, track, and bridge further complicates the dynamic loading scenario. (b) Non-unique solutions and false positives: solving the optimization problem for damage identification often results in non-unique solutions, where multiple combinations of damaged members and damage intensities produce similar system responses, creating ambiguity. This issue is exacerbated by the limited number of sensors and measurement noise, making it challenging to accurately determine the location and severity of damage [25]. False-positive damage detection, where the algorithm incorrectly identifies damage in healthy members, can lead to unnecessary maintenance efforts, increasing costs and causing disruptions to railroad operations [26]. (c) Lack of frameworks for multimodal data integration: effective frameworks for integrating multimodal data from various sensors and incorporating prior information about the bridge's state into the updating process are lacking. The integration of diverse data streams, such as strain, acceleration, and displacement measurements, along with inspection reports and maintenance records, is crucial for accurate damage detection and localization [27]. However, existing frameworks are not well-suited for combining this heterogeneous information, limiting their real-world applicability. Developing a comprehensive framework that can efficiently handle multimodal data and leverage prior information is needed for advancing damage identification. (d) Limited access and data collection challenges: railroad bridges are often in remote locations, making the collection and transmission of sensor data challenging [11]. Harsh environmental conditions and the need to minimize disruptions to railroad operations further hinder data acquisition. Consequently, data-driven approaches must be designed to perform effectively with limited training data.

To address these gaps, a physics-informed neural network (PINN) approach for damage identification of truss railroad bridges is proposed. By incorporating governing physical equations into the neural network's learning process, PINNs can ensure that models better represent the underlying physics [28,29], allowing for more accurate predictions of complex system behaviors, even with limited or noisy data [30,31]. Herein, the second-order ordinary differential equations (ODEs) governing the dynamics of the train-bridge interaction is directly encoded into a recurrent neural network (RNN) based deep learning architecture. The proposed unsupervised learning approach aims to identify damaged members and update the model to reflect the current condition of the railroad bridge. The PINN-based model uses the moving train load time history and measured output responses as inputs for damage quantification and localization, learning damage parameters from unlabeled data during training without explicit labels. In this formulation, unknown parameters are iteratively updated to minimize the discrepancy between predicted and observed responses. Furthermore, the proposed updating pipeline is designed to seamlessly integrate prior knowledge from inspections and drone surveys, enabling context-aware updating and assessment of bridge's condition. The Calumet Bridge, a steel truss railroad bridge located in Chicago, Illinois, is used to validate the proposed strategy; various damage scenarios, along with the in-service train loads, are simulated using Python. The performance of the proposed PINN-based strategy on the Calumet Bridge demonstrates its effectiveness in accurately identifying damage, while maintaining low false-positive rates, even in the presence of noisy data.

The main contributions of this study include: (a) an unsupervised approach to damage identification in truss railroad bridges using PINNs, (b) integration of prior information from multimodal sources into the deep learning-based updating pipeline, and (c) application of PINNs for damage identification of large multi-degree-of-freedom (MDOF) and linear time-varying (LTV) structural systems, validated through simulations on a full-scale truss railroad bridge. By embedding physics-informed kernels within the updating routine, this approach aims to enable more accurate damage identification and bridge model updating. This approach potentially addresses some of the unique challenges in railroad bridge assessment to enhance structural health monitoring and maintenance strategies.

## 2. BACKGROUND

To provide context to the research reported herein, this section briefly describes the two main approaches that have gained prominence in the application of PINNs for structural dynamics problems.

The first approach uses deep neural networks to approximate solutions for differential equations, with the optimization process guided by a loss function that adjusts the hyperparameters to satisfy initial and boundary conditions, as well as the governing equations, through collocation points [29,32]. PINNs have been applied to different problems in structural dynamics, such as input identification [33], vibration analysis [34], and system identification [35]. For example, Zhang et al. [36] used a physics-informed variational autoencoder architecture to identify the structural excitation in a system using responses measured from ambient vibration. In this study, the authors validate the approach using only an LTI structural system, specifically a building frame model. Liu and Meidani [35] propose PIDynNet, a supervised learning approach using fully-connected neural networks (FCNNs) with physics-based loss terms for parameter estimation of nonlinear structural systems. Chen et al. [34] improve the standard PINN formulation for long-duration structural vibration problems by employing a time-marching scheme. Yamaguchi and Mizutani [37] use a

PINN with a supervised FCNN architecture for damage identification in RC bridge piers, employing a nonlinear MDOF reduced-order model of the bridge pier. Most studies have been limited to small-scale, simplified systems with few degrees of freedom, which restrict their applicability to more complex, real-world structures. While these approaches demonstrate effectiveness in supervised learning applications, they fall short in addressing the challenges posed by limited labeled training data.

The second approach focuses on constructing hybrid models that integrate reduced-order physics-informed models within deep neural networks [38–40]. In this approach, the computational cost of physics-informed kernels is kept comparable to the linear algebra operations typical in neural network architectures. The backpropagation algorithm used to update unknown parameters within the neural network depends on adjoint calculations, which can be performed using automatic differentiation [41]. This requirement means that the physics kernel must allow gradient flow with respect to unknown parameters. Renato et al. [38] proposed a method to identify unknown damping parameters in structural systems by embedding a fourth-order RK integrator-based multi-degree-of-freedom (MDOF) differential equation solver into a physics-informed recurrent neural network (RNN) cell. The researchers used RNNs to capture temporal dependencies in the structural system, while the physics-based solver ensured adherence to the governing equations of motion. However, this method is unable to perform damage identification for railroad bridges while accounting for the LTV nature of the system and bridge-train interaction. Thus, an extension of this approach is developed herein, the details of which are discussed in the following section.

## 3. PROPOSED APPROACH

This section provides an overview of the proposed PINN-based damage identification approach for truss railroad bridges, as shown in Figure 3.1. The green box in this figure represents the forward pass in the PINN-based architecture, illustrating the first step. The LTV structure is modeled with a Phy-RNN cell (blue box) for which the loading time history is taken as input. The output is then the predicted response time history.

Subsequently, this section describes the damage simulation approach, and the model update step where an unsupervised learning scheme is adopted for identification of damage on truss railroad bridges (see red bounding box region in Figure 3.1). Each of these steps are described in the remainder of this section.

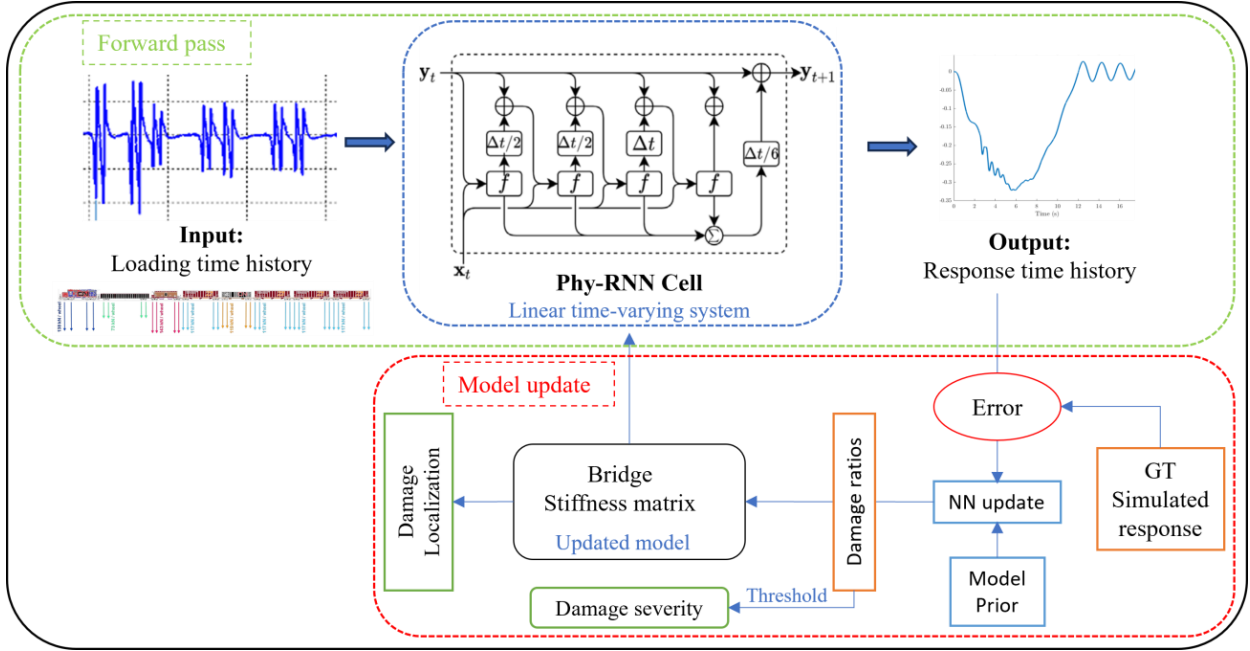

**Figure 3.1: Workflow for damage identification in railroad bridges using unsupervised PINN.**

## 3.1. PINN-based architecture forward pass

The proposed PINN-based architecture to perform damage identification for LTV structural systems is described in this subsection. The proposed architecture explicitly embeds a reduced-order physics model into the deep learning pipeline, enabling the integration of domain knowledge and data-driven learning. At the core of the proposed PINN-based architecture is a recurrent neural network (RNN) cell [42] that is designed to perform numerical integration. Building upon the model proposed by Nascimento et al. [38] for solving ordinary differential equations (ODEs), a PINN-based architecture is designed to solve the governing ODE of a linear time-varying (LTV) railroad bridge-train system subjected to moving train masses. In a conventional RNN cell (see (a)), for every timestep $t$, the next state ($y_t$) is predicted based on the current input ($x_t$) and the state in previous timestep ($y_{t-1}$) by applying a transformation $f(.)$ as described by the equation $y_t = f(y_{t-1}, x_t)$. The RNN cell proposed here, referred to as Phy-RNN (see Figure 3.2(b)), performs numerical integration for a railroad bridge-train system using data from train crossing events. Within the Phy-RNN cell, an explicit fourth-order Runge-Kutta (RK-4) based integrator is designed in Python to solve the ODE governing the bridge-train dynamics. Critical to this approach is the ability to deal with time-varying systems such as is seen during train crossings for a railroad bridge-train system. The design of the RK-4 integrator using the PyTorch library [43] take into account two important considerations: (a) the computations approach linear algebra time complexity [44], which help maintain reasonable training times, and (b) the gradients with respect to the trainable unknown parameters exist to facilitate backpropagation-based optimization. While the implemented RK-4 integrator herein is computationally efficient per timestep, this solver is conditionally stable and necessitates very small timesteps to maintain stability. Therefore, for handling more complex bridge systems with higher frequency content, an implicit RK integrator is also designed herein to accommodate larger timesteps. The design of the implicit RK integrator to solve the LTV bridge-train system is discussed in Section 4.5.

Primarily, the Phy-RNN cell processes three inputs: (a) the train loading time history vector of size $m$, where $m$ represents the number of free degrees of freedom (DOFs) in the structural system, (b) the initial or previous state of the system, sized $2 \times m$ at timestep $(t-1)$, and (c) the transient mass matrix of size $m \times m$. The output of the Phy-RNN cell is the predicted state of the bridge-train system, sized $2 \times m$ at timestep $t$. By incorporating the time-varying mass matrix computed at each timestep, the Phy-RNN cell effectively models the LTV dynamics of the bridge-train system.

The next subsection describes the damage simulation approach, and the unsupervised learning scheme used in this research for damage identification on truss railroad bridges.

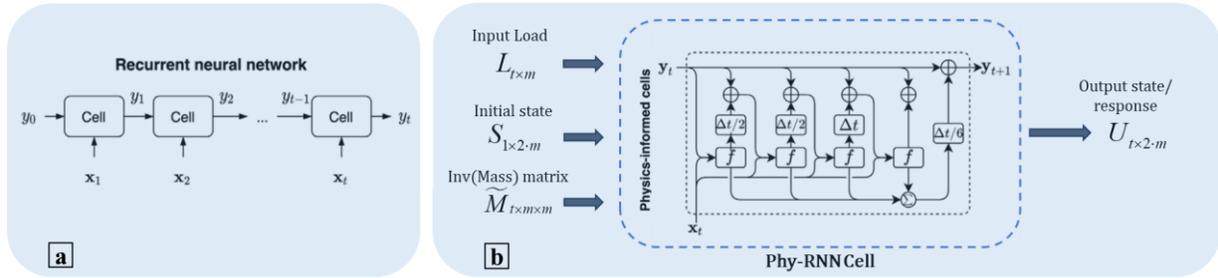

**Figure 3.2: Recurrent neural network cell: (a) Standard RNN cell, (b) Phy-RNN cell with RK-4 based bridge-train system ODE solver.**

## 3.2. Damage simulation and unsupervised learning scheme

This section first describes how damage for the bridge-train system is modeled; subsequently, the associated model update step and the unsupervised learning scheme is detailed, as illustrated in the red bounding region of Figure 3.1. Data from various bridge damage scenarios is typically required to train a supervised deep-learning model; however, obtaining sufficient training data from a damaged structure, such as a railroad bridge with known damage levels, is often challenging and impractical in real-world scenarios. To address this limitation, this research relies on simulating different damage scenarios on the finite element (FE) model of a railroad bridge, and employs an unsupervised learning scheme. This study focuses on identifying damage in structural members of truss railroad bridges, typically modeled as a reduction in member stiffness [17], which can be due to section loss, excessive corrosion, fatigue and buckling. To simulate various damage scenarios, different damage levels are introduced to selected structural members of the truss railroad bridge by assigning deviation ratios in the range 0 to 1.0, where 1.0 represents an undamaged state and 0 indicates complete stiffness loss. Notably, deviation ratios greater than 1.0 are also considered to simulate members with higher-than-assumed stiffness; evaluating the proposed approach's ability to detect underestimated stiffness in structural members. This study assumes that the input train loading time history is available for train crossing events; that can be measured on-site using strain gauges mounted on the railroad track [11]. To serve as a substitute for measured responses on a bridge with a known damaged state, train-wheel configurations at velocities of interest are simulated on the FE model to generate ground-truth output responses. Several cases with random deviation ratios assigned to structural members at different locations of a railroad bridge are considered to evaluate the performance of the proposed PINN-based strategy.

The model update step in this PINN-based unsupervised learning scheme then begins with a surrogate model of the bridge, initially assumed to be in a healthy state. This initial FE model is typically developed from the bridge's design drawings. In this baseline model, all structural members start with a deviation ratio of 1.0, representing an undamaged condition. In the forward pass of this unsupervised learning scheme, member-level stiffness matrices are updated according to the deviation ratios, and the global stiffness matrix is reassembled using the bridge's connectivity matrix. Assuming constant train velocity during bridge crossing, the inverse of the transient mass matrix at each timestep, $t$, can be precomputed before the learning routine starts, reducing computation time during the forward pass. The Phy-RNN cell then predicts the system response based on the updated global stiffness matrix. In the global stiffness matrix update step, each local member stiffness matrix is scaled by its respective deviation ratio ($k_i^d$) of that member, where $k_i^d$ is the learnable deviation ratio parameter for the $i^{th}$ structural member. These updated member stiffness matrices are then assembled into the global stiffness matrix, maintaining its inherent symmetry to prevent system instability. This approach reduces the number of learnable parameters to the number of structural members, avoiding the need to directly update the global stiffness matrix, which would require $n \cdot (n+1)/2$ unique unknown parameters, where $n$ is the size of the global stiffness matrix. Directly updating the global matrix would not only expand the parameter search space but also lead to risk of instability from an ill-conditioned stiffness matrix. Thus, the local-to-global update strategy ensures stability, reduces the parameter space, and improves physical interpretability by facilitating the localization of damaged members.

The discrepancy between the predicted response of the initial healthy bridge model and the simulated ground-truth response of the damaged bridge guides the model optimizer in updating neural network parameters. In addition, the influence of available prior information on the railroad bridge is incorporated into the network parameters before the optimizer update step, as will be described in detail in Section 4.6. During the optimization step, the model adjusts the unknown deviation ratios associated with the structural members of the truss railroad bridge. Through this unsupervised learning approach, the proposed strategy aims to identify damaged members and quantify their damage severities. The efficient implementation of the proposed PINN-based approach described in this section will enable damage identification of large MDOF linear time-varying systems. The next section details the modeling of the truss railroad bridge-train system to validate the proposed approach for identifying and localizing damage.

## 4. RAILROAD BRIDGE SYSTEM MODELING

The physics-based modeling used to capture the dynamics of the truss railroad bridge considering the bridge-train interaction problem is the focus of this section. The railroad bridge selected to validate this damage identification study is an open-deck steel truss bridge spanning the Calumet River near Chicago, Illinois (see Figure 4.1, highlighted in red with 'CN Bridge'). Opened for service in 1971, the Calumet bridge is approximately 95 meters long, 21 meters tall, and 10 meters wide, with a truss structure comprised of a series of interconnected steel members. First, the development of the FE model for the Calumet bridge is described, along with the formulation of the governing equations of motion for the linear time-varying nature of the system. Although the Calumet bridge has two tracks, a simplified 2D model of the bridge is developed for the initial study which neglects the asymmetric loading of the bridge. Subsequently, a 3D model of the Calumet bridge is developed that accommodates the two tracks on the bridge. Further, to increase computational efficiency, a reduced-order 3D model is developed. For performing damage identification on the 3D model efficiently an implicit RK integrator is designed that handles the LTV bridge-train system. Finally, this section describes the strategy used in this research to incorporate available prior information about the bridge model into the PINN-based damage identification and updating routine.

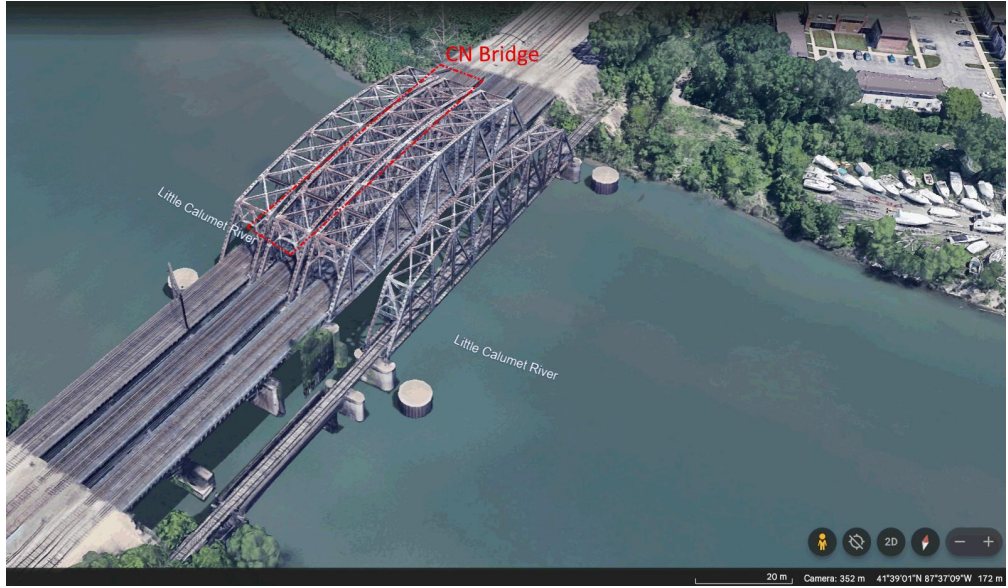

**Figure 4.1: Calumet railroad bridge near Chicago, IL. Image source: Google Earth [45].**

## 4.1. Hybrid bridge model

A brief overview of the development of a hybrid model for the Calumet bridge is presented in this subsection. Kim et al. [46] developed a hybrid model of the Calumet Bridge with a simplified train-track-bridge interaction formulation to estimate the dynamic responses for model updating. The formulation employed herein builds on this work. For the convenience of the reader, a brief overview of the modeling is presented; for detailed information, interested readers can refer to Kim et al. [47]. In this research, the entire modeling and FE simulation are implemented in Python to integrate seamlessly with the PINN-based deep learning pipeline developed on Python.

*4.1.1 Bridge and Track Modeling*

Considering first only the railroad bridge (excluding the train), the EOM can be represented as a linear time-invariant system given by:

$$M_B \ddot{u}_B(t) + C_B \dot{u}_B(t) + K_B u_B(t) = f_B(t) \tag{1}$$

where $M_B, C_B$, and $K_B$ are the mass, damping, and stiffness matrices of the bridge, respectively; $u_B(t)$ is the displacement vector for the structure, and $f_B(t)$ is the vector of applied forces. The track is modeled as a continuous beam using the assumed modes method. A lumped moving mass model is used to simulate train cars crossing the bridge. The resulting equation for the train-track system is given by:

$$(M_R + \Delta M_R(t))\ddot{q}_R(t) + C_R \dot{q}_R(t) + K_R q_R(t) = p_R(t) \tag{2}$$

where $M_R, C_R,$ and $K_R$ are the mass, damping, and stiffness matrices of the rail, respectively; $\Delta M_R(t)$ is a time-varying mass matrix corresponding to the moving train masses, $q_R(t)$ is the generalized displacement of the rails, and $p_R(t)$ is the external force vector. The next subsection describes how to combine the bridge and train/track system to obtain the EOM of the LTV system.

*4.1.2 Bridge-Train-Track Interaction*

To couple the bridge and train-track models, an interaction layer comprised of a discrete spring-dashpot system is introduced to represent the sleepers or rail ties [47]. Then, the combined hybrid system of equations can be obtained as:

$$M_{total}(t)\ddot{u}(t) + C_{total}\dot{u}(t) + K_{total}u(t) = p(t) \tag{3}$$

where $M_{total}(t), C_{total},$ and $K_{total}$ are the mass, damping, and stiffness matrices of the augmented coupled system, respectively; $u(t) = \{q_R; u_B\}$ is the total displacement vector, and $p(t) = \{p_R; f_B\}$ is the total force vector. Herein, Equation (3) contains coupled terms that introduce interactions between the rails and the bridge structural system. Based on this formulation, the simplified 2D model developed for the Calumet bridge with a single-track configuration is discussed next.

**4.2. Calumet bridge – 2D model**

This subsection describes the formulation of the Calumet bridge 2D model, illustrated in Figure 4.2, developed in Python for the initial study. The 2D model is used to facilitate preliminary analysis and validate the effectiveness of the PINN based damage identification approach before extending it to the more complex 3D model. The 2D formulation uses plane frame elements for modeling bridge structural members, and continuous simple beam elements to represent the rails. This model consists of 37 structural members and 20 nodes, each with two DOFs after condensing out the rotational DOFs through static condensation. The rails are modeled using the assumed modes method, where the number of modes to be used was determined based on the calibration study by Kim et al. [10]. This study demonstrated that five modes were sufficient to achieve reasonable correlation between the estimated global responses and site-measured data for the Calumet bridge. Next, the 3D model developed for the Calumet bridge to account for torsional effects from asymmetric loadings is described.

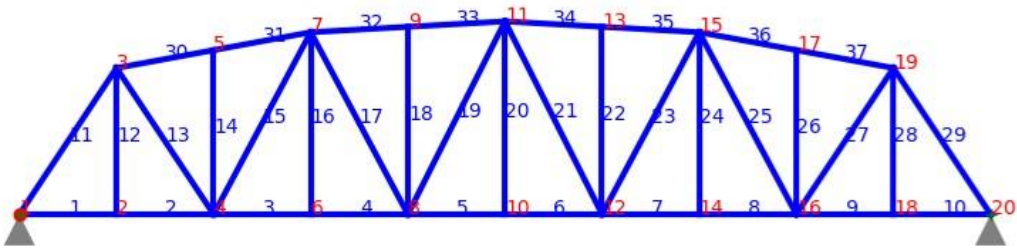

**Figure 4.2: Calumet bridge 2D model in python**

### 4.3. Calumet bridge – 3D model

The 3D Calumet bridge model is built in Python, including the double-tracks that allow for cases of asymmetrical train loads; this section describes this model. The initial 3D bridge model (illustrated in Figure 4.3(a)) representing the undamaged state of the bridge, with system matrices, $M_B, C_B$, and $K_B$, is constructed based on the calibrated FE model developed by Kim et al. [10]. The cross-bracing at the top of the bridge has a relatively small cross-sectional area compared to the other structural members and has less influence on the global response of the bridge due to vertical train loads. For simplification, this cross-bracing is excluded from the members considered in the damage identification study. After condensing out the rotational degrees of freedom, the 3D model developed herein consists of 94 nodes with 252 free DOFs; for the damage identification study, a total of 157 structural members (including the green-colored cross bracings at the bottom, in Figure 4.3(a)) are considered.

As described in Section 3.1, while utilizing the explicit RK-4 based integrator design within the Phy-RNN cell for the forward pass, requires an appropriately small timestep size ($dt$) to achieve a stable and accurate simulation of the structural system. Typically, for structural systems, the timestep must be small enough to capture the highest natural frequency of the system. As a rule of thumb, the timestep is selected as $dt \leq T_{min}/10$, where $T_{min} = 1/f_{max}$, and $f_{max}$ is the highest significant natural frequency of the system. For the 3D Calumet bridge model, the highest natural frequency is 518.8 Hz, necessitating a timestep of at least 0.0002 seconds, which increases the computational demands. To efficiently perform damage identification on a system with large number of DOFs, Guyan reduction is employed herein to derive a reduced-order model, as detailed below.

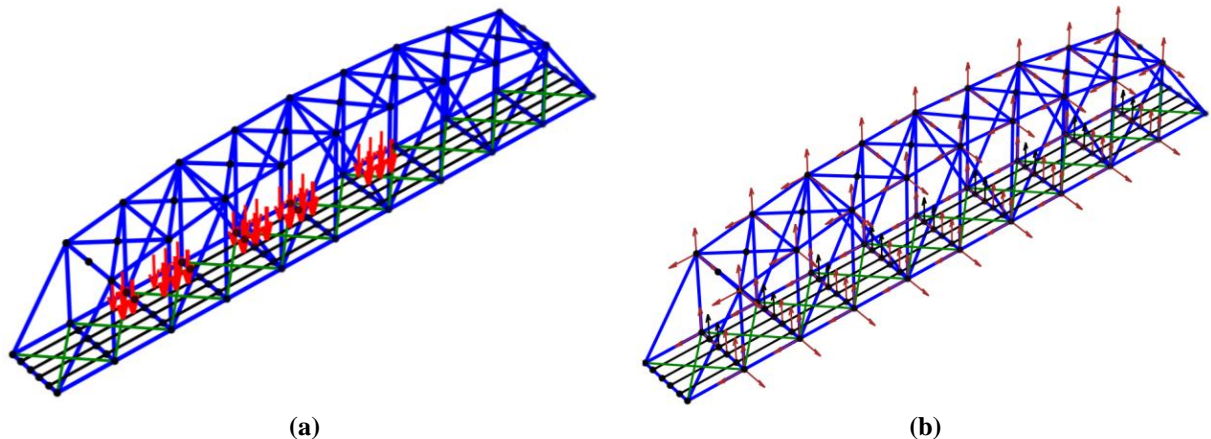

(a)          (b)

**Figure 4.3: Calumet bridge: (a) 3D model in Python with moving train load, and (b) Guyan reduced-order model with selected master degrees of freedom**

### 4.4. Guyan reduced model

Guyan reduction is a static condensation approach for obtaining reduced-order model for linear time-invariant (LTI) structural systems [48]. As the bridge-train system in this research is an LTV system, Guyan reduction cannot be directly applied. A simple yet effective approach is proposed herein to obtain the reduced-order model of the LTV bridge-train system. First, considering the bridge and train-track system separately, Guyan reduction is applied only on the LTI bridge structural system. Then, the obtained reduced-order bridge model is combined with the track-train system as in Equation (3) to obtain the augmented LTV reduced-order structural system. Note that Guyan reduction requires choosing master and slave DOFs in the structure. Herein, the train loading is applied to the bridge via the stringer beams, which are modeled to be in contact with the bridge nodes at the rail level for load-transfer. So, the DOFs of interest on the nodes at rail level of the bridge ($u_B^{rail}$), shown as black colored arrows in Figure 4.3(b), are selected to be among the master DOFs [49]. In addition, the master DOFs include the bridge's DOFs of interest where responses are measured ($u_B^{add}$), shown as brown colored arrows in Figure 4.3(b). Increasing the number of master DOFs introduces a trade-off: it improves the accuracy of simulating the full 3D model's responses, particularly at the master DOFs, but also increases the complexity and size of the reduced-order model. The master DOFs selected for this study are illustrated in Figure 4.3(b). Herein, for the reduced-order model the retained master DOFs, $u_B^G = \{u_B^{rail}; u_B^{add}\}$, are related to the bridge's DOFs ($u_B$) as $u_B = T u_B^G$, where

$$T = \begin{bmatrix} I_{n_m \times n_m} \\ -K_{B22}^{-1} K_{B21} \end{bmatrix} \quad (4)$$

$n_m$ is the number of retained master DOFs, and the bridge stiffness matrix ($K_B$) is partitioned into blocks and permuted based on DOF numbering, with subscripts 1 and 2 representing the master and slave DOFs, respectively. Then, the Guyan reduced system matrices are obtained as

$$\begin{aligned} M_R^G &= T^T M_B T \\ K_R^G &= T^T K_B T \\ C_R^G &= T^T C_B T \end{aligned} \quad (5)$$

Using Equation (3), the reduced system matrices are combined with the track-train system to obtain the combined Guyan reduced bridge-train-track system matrices as $M_B^G, K_B^G$, and $C_B^G$. For the 3D Calumet bridge model, the reduced-order model obtained here has system matrices of size 136 x 136, and a highest natural frequency of 131.37 Hz, which is nearly 4 times smaller than the highest natural frequency for the full 3D model. Herein, to be able to use larger timesteps for more efficient computation while maintaining stability and accuracy, an implicit RK integrator is designed as described below.

### 4.5. Implicit RK integrator

The implicit RK integrator employed in this study is based on the Radau IIA solver with two stages and third-order accuracy [50]. The algorithm designed for solving the Guyan reduced bridge-train system using the implicit RK formulation is illustrated in Figure 4.4. Radau IIA is both A-stable and L-stable, enabling the use of larger timesteps without numerical instability. The Butcher coefficients [50] used for the third-order method are shown in Step 1 of Figure 4.4. Herein, for computational efficiency, the external force is assumed to remain approximately constant during the timestep $dt$, allowing the $c$ term of the Butcher coefficients to be omitted. The system state is updated by solving the implicit equations in block-linear form with PyTorch, shown in Step 3 of the algorithm.

The initial state of the system at $t = 0$ is set within the Phy-RNN cell as $X_0 = \mathbf{0}_{2 \times N_g}$ where $N_g$ represents the number of DOFs in the Guyan reduced system. The developed algorithm for the Phy-RNN cell, which utilizes the Guyan reduced bridge-train system for learning damage parameters on the full 3D bridge model, is presented in Figure 4.5. Herein, the operations involved in updating the system's state are designed to allow flow of gradients, enabling parameter updates with backpropagation. As illustrated in Step 2 of the algorithm, Guyan reduction is performed only once for every batch using the updated bridge stiffness matrix. To estimate the time-varying mass matrix at each timestep, the JAX-accelerated python library is used with a JIT function [51]. This function performs the required matrix operations with an optimized static computational graph, providing over tenfold speed improvement compared to the standard implementation. The next state for each timestep in a batch is estimated using the implicit RK integrator described in Figure 4.4. Using this implicit RK integrator along with the reduced-order model, the next subsection outlines the strategy for incorporating prior information about the railroad bridge into the damage identification pipeline.

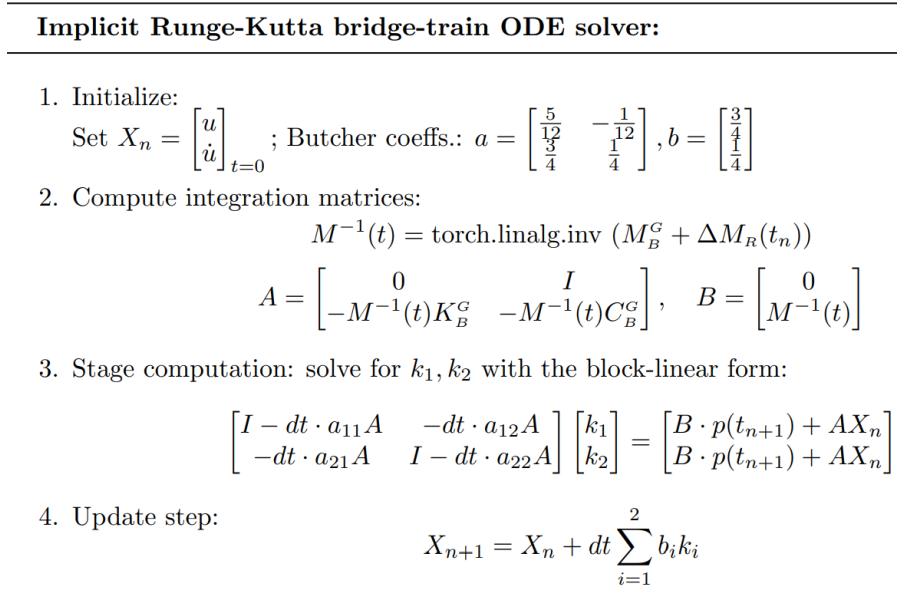

**Implicit Runge-Kutta bridge-train ODE solver:**

1. Initialize:
   Set $X_n = \begin{bmatrix} u \\ \dot{u} \end{bmatrix}_{t=0}$ ; Butcher coeffs.: $a = \begin{bmatrix} \frac{5}{12} & -\frac{1}{12} \\ \frac{3}{4} & \frac{1}{4} \end{bmatrix}, b = \begin{bmatrix} \frac{3}{4} \\ \frac{1}{4} \end{bmatrix}$

2. Compute integration matrices:
$$M^{-1}(t) = \text{torch.linalg.inv}\left(M_B^G + \Delta M_R(t_n)\right)$$
$$A = \begin{bmatrix} 0 & I \\ -M^{-1}(t)K_B^G & -M^{-1}(t)C_B^G \end{bmatrix}, \quad B = \begin{bmatrix} 0 \\ M^{-1}(t) \end{bmatrix}$$

3. Stage computation: solve for $k_1, k_2$ with the block-linear form:
$$\begin{bmatrix} I - dt \cdot a_{11} A & -dt \cdot a_{12} A \\ -dt \cdot a_{21} A & I - dt \cdot a_{22} A \end{bmatrix} \begin{bmatrix} k_1 \\ k_2 \end{bmatrix} = \begin{bmatrix} B \cdot p(t_{n+1}) + A X_n \\ B \cdot p(t_{n+1}) + A X_n \end{bmatrix}$$

4. Update step:
$$X_{n+1} = X_n + dt \sum_{i=1}^{2} b_i k_i$$

**Figure 4.4: Algorithm with pseudocode for implicit Runge-Kutta Radau IIA [50] solver for bridge-train dynamics.**

```
Phy-RNN cell with Guyan reduction:

1. Initialize:
        /* Last step from previous batch or None */
        X_n ← Initial_state
2. Update Guyan reduced system:
        /* Updated bridge stiffness matrix with damage */
        K_d ← Bridge_stiffness_damage(k^d)
        /* PyTorch Gradient flow */
        K_B^G, M_B^G, C_B^G ← Guyan_reduction(K_d)
3. Estimate the Updated state:
        /* Loop for time-stepping */
        for t in (n, batch_size):
                /* JAX accelerated functions */
                ΔM_R(t_n) ← Time_varying_mass(t)
                p(t_n) ← External_force(t)
                X_{n+1} ← Implicit_RK(p(t_n), ΔM_R(t_n), X_n)
```

**Figure 4.5: Algorithm for Phy-RNN cell with Guyan reduced system.**

## 4.6. Incorporation of multimodal prior information

Context-aware bridge damage assessment refers to integrating factors such as environmental conditions, operational factors, and available prior knowledge about the bridge from diverse sources into the assessment process. Such prior knowledge for railroad bridge assessment can be derived from multimodal sources, including visual inspections conducted by bridge engineers, drone surveys providing visual indicators of damage on structural members, and sensitivity analysis of members using finite element (FE) models. In this study, the learning scheme incorporates prior knowledge in two ways: (i) *modifying the initial deviation ratio*: instead of initializing the optimizer with $k_i^d = 1.0$ (healthy state) for all structural members, the initial value of $k_i^d$ is modified for members that are likely damaged based on prior information. Herein, a heuristic approach of weighted combination [52] is used to account for both the prior estimated damage and the confidence level associated with that estimate. The initial deviation ratio is computed as:

$$k_i^{d_{\text{initial}}} = (1-p) \times k_i^{d_{\text{healthy}}} + p \times k_i^{d_{\text{prior}}} \qquad (6)$$

where, $k_i^{d_{\text{initial}}}$ is the initial deviation ratio for $i^{th}$ member, $p$ is the confidence level (in the range 0-1.0) associated with the prior estimate, $k_i^{d_{\text{healthy}}}$ represents the healthy (undamaged) state, and $k_i^{d_{\text{prior}}}$ is the prior estimate of the deviation ratio. For instance, if a bridge inspector, based on site conditions, asserts that a member has degraded by 20% with a confidence level of 70%, from Equation (6) it can be estimated that $k_i^{d_{\text{initial}}} = 0.86$. (ii) *gradient scaling*: a scaling factor is applied to adjust the gradients estimated by backpropagation for the corresponding structural members. Herein, the structural members that are identified as less sensitive to measured responses from the

FE model are assigned higher scaling factors. By amplifying the gradients of these less sensitive parameters, the optimizer can take larger steps in updating them, potentially improving the convergence speed and the parameter estimation accuracy. Mathematically, the modified gradient is estimated as $\nabla^i_{mod} = s_i \times \nabla^i$, where $s_i$ is the scaling factor, and $\nabla^i$ represents the original gradient for the member. As will be shown in the next section, these methods effectively integrate expert knowledge and multimodal information into the learning process, steering the model towards identifying the most likely damaged members.

## 5. RESULTS

This section validates the efficacy of the proposed PINN-based damage identification approach through several damage scenarios on the Calumet bridge under simulated train crossing events. First, the training configuration utilized for the unsupervised learning scheme is described, including the details of the loss function employed for optimization. Next, the model's performance is assessed on the 2D Calumet bridge model across various damage cases. Finally, the proposed PINN based strategy is evaluated for simulated damage scenarios in the 3D Calumet bridge model to demonstrate its effectiveness on a large multi-DOF bridge-train system.

### 5.1. Training configuration and loss function

First, details of the hyperparameter configuration used for the unsupervised learning scheme are discussed. For the Phy-RNN cell, each epoch corresponds to a full pass of the simulated time-history response during a train crossing event. A batch-wise training approach is employed to update the model's learnable parameters in batches, thereby reducing computational costs. Based on trial-and-error, a batch size of 4 to 8 was found to perform well. In the batch-wise approach, the first batch assumes an initial state of zero (before the train enters the bridge). For subsequent batches, the initial state is updated using the predicted state from the last timestep of the preceding batch. To enhance the robustness of the unsupervised learning approach, particularly in the absence of explicit labels for damaged members, a physical constraint is introduced during parameter updating. This constraint ensures that the minimum value of the deviation ratio remains at 0.01, preventing structural instability. During each epoch, the PINN based model processes the input load time history in batches and predicts the corresponding state. Notably, damage identification is performed using data from a single train-crossing event. The predicted response is compared with the simulated ground-truth response for the damage case to compute the loss, which guides the optimizer in updating the model parameters.

The choice of loss function and optimizer plays a crucial role in the training process and can significantly impact the model's performance. The Phy-RNN model outputs the structural response for a batch of input load time history, given as a state vector with displacement and velocity response of the free DOFs. For the case when acceleration responses are also used for damage identification, the acceleration vector is estimated with the finite difference method by approximating it as the first-order derivative of the velocity vector. Different loss metrics were experimented, the mean-absolute error (MAE) or L1 norm, mean-squared error (MSE) or L2 norm, and a combination of L1 and L2 norm called SmoothL1Loss (also known as Huber loss) [53]. Overall, the best results were obtained for SmoothL1Loss given as

$$l_n = \begin{cases} 0.5(x_n - y_n)^2 / \beta, & \text{if } |x_n - y_n| < \beta \\ |x_n - y_n| - 0.5\beta, & \text{otherwise} \end{cases} \quad (7)$$

where $l_n$ is the Huber loss for the $n^{th}$ sample, $x^n$ is the predicted value, $y^n$ is the true value, and $\beta$ is a hyperparameter that controls the transition point between the quadratic and linear portions of the loss.

To incorporate different types of measured responses on a bridge, e.g., displacement and acceleration time-histories, a combined loss function is utilized. As damage in structural members has different sensitivity to displacement and acceleration response of the bridge, appropriate regularization constants are used. Through trial and error, regularization weights of 0.9 and 0.1 were assigned to the displacement ($l_n^{displacement}$) and acceleration ($l_n^{acceleration}$) Huber losses, to achieve a weighted combined loss of

$$l_n^{combine} = 0.9 \times l_n^{displacement} + 0.1 \times l_n^{acceleration} \tag{8}$$

This weighted loss function balances the respective contributions to the overall loss attained, yielding good damage identification results. In this way the proposed approach can use different types of available data for improved optimization.

To better stabilize the learning process, the Cyclic Learning Rate (CyclicLR) scheduler from PyTorch [54] is employed. Learning rate scheduling can be used to dynamically adjust the learning rate during training. This adaptive learning rate method varies the learning rate between a lower boundary and an upper boundary over a cycle of a certain number of iterations. The exponential range scheduler [55] is configured, allowing for a gradual increase and decrease in learning rates. This approach aims to balance the trade-off between faster convergence and avoiding local minima, potentially leading to improved model performance.

To optimize the model parameters and minimize the loss function, several optimization algorithms are explored, including Stochastic Gradient Descent (SGD), AdamW, and RMSprop. Damage in different structural members of a bridge exhibits varying levels of sensitivity to measured responses. Optimizers capable of dynamically adjusting learning rates for each learnable damage parameter were observed to yield superior performance. Through extensive hyperparameter tuning, AdamW and RMSprop emerged as particularly effective optimizers. The model's performance is evaluated first for the 2D Calumet bridge case and subsequently for the 3D case for different damage scenarios.

### 5.2. Results for the 2D Calumet bridge model

First, a scenario involving damage in three structural members of the 2D Calumet bridge model is discussed. Then, the approach used in this research to minimize occurrence of false-positives in damage identification is detailed. Following evaluation of the model's performance in the presence of noisy measurements, quantitative evaluation is performed across various damage cases.

*5.2.1 Case Study: Three damaged members*

In this case study, damage is simulated in three structural members of the 2D Calumet bridge model. To evaluate the damage identification process for different types of members, the selected member numbers are: 5 (horizontal), 17 (diagonal), and vertical (20), as shown in red in Figure 5.1. The damage intensities for these members are set to: 0.7, 1.25, and 0.8, respectively. A deviation ratio less than 1.0 indicates a reduction in the member's structural stiffness, while a ratio greater than 1.0 suggests that the actual member stiffness is higher than anticipated for the base model. The AdamW optimizer is employed here with a learning rate of 0.01. The loss metric used is the Huber loss of the displacement response at the free DOFs. In this study, a train running at 50 MPH is simulated with train configuration and axle-loads based on the campaign monitoring conducted by Kim at al. [46]. The train crossing event duration is approximately 10 secs; thus, the input time history and output response consist of about 23,800 timesteps when simulated using a timestep of 0.0004 sec. The dataset is split into 4 batches for training, with a batch size of 5950. Herein, the explicit RK-4 integrator is used within the Phy-RNN cell for performing damage identification. The evolution of the learned deviation ratios for the 37 structural members during the learning process is visualized in Figure 5.2 (a). Convergence is achieved within 200 epochs, as shown by the flattening loss curve in Figure 5.2(b), with deviation ratios for the three structural members approaching ground truth values in Figure 5.2(a).

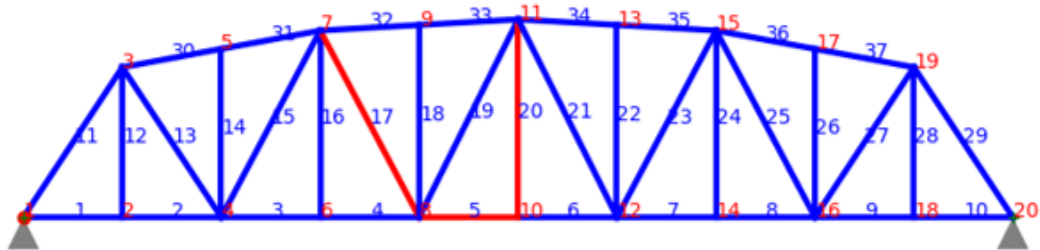

**Figure 5.1 Simulated case on 2D Calumet bridge with damage in three structural members (shown in red): members numbers - 5, 17, and 20.**

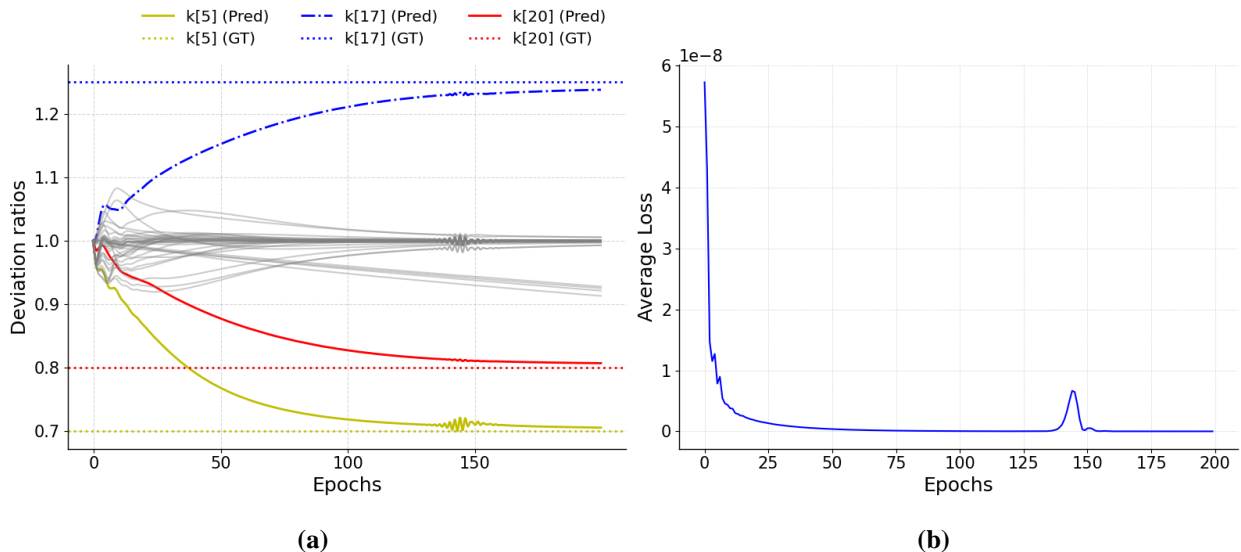

**Figure 5.2 Training progress for the case with three damaged members using AdamW: (a) evolution of deviation ratios for the 37 structural members, (b) training loss curve.**

Herein, a member is assumed to be classified as a false positive for damage if the error tolerance is more than 5% for the undamaged members. The PINN based model accurately identifies the three damaged members, with predicted deviation ratios as [0.706, 1.238, 0.807] for the members [5, 17, 20], as shown in Figure 5.3 (a). An average damage identification accuracy of 99.10% is obtained with a maximum error of 0.948% for the damaged members. In Figure 5.2(a), the gray-colored lines represent the evolution of deviation ratios for the undamaged members in the bridge. The error bar plot in Figure 5.3(b) shows that there are four members classified as false positives with a maximum error of 8.67%. Using a different set of hyperparameters, combined with prior information about the 2D bridge model, helps reduce false positives, as detailed in the following subsection.

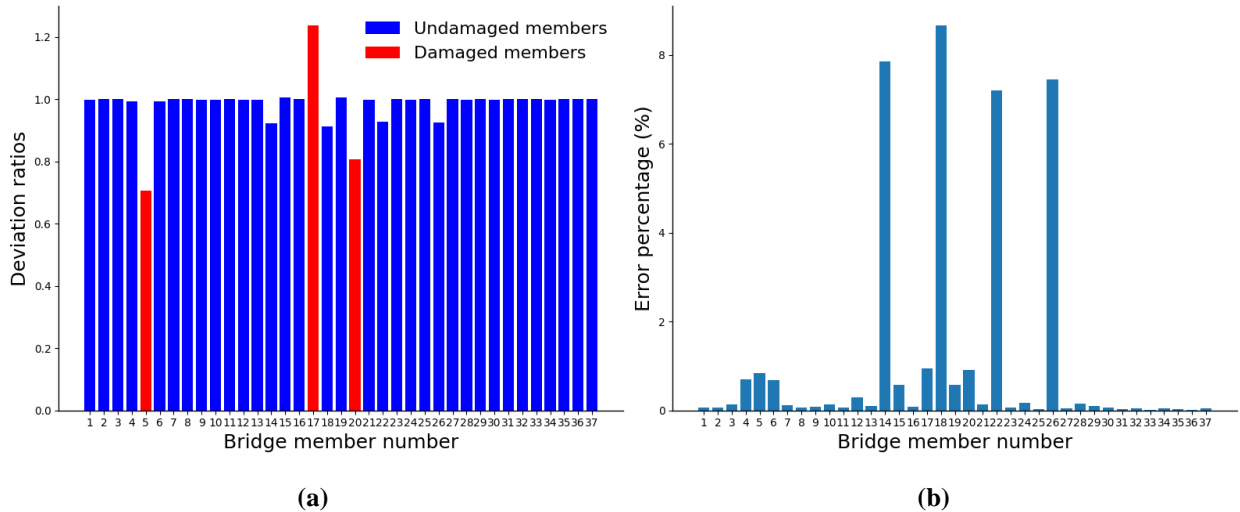

**Figure 5.3 Bar plots for Calumet 2D bridge model case study: (a) updated deviation ratios with three damaged structural members (shown in red), (b) percentage error in deviation ratios.**

*5.2.2 Minimizing false positives*

In the results shown in Figure 5.3, the false-positive results were obtained for members: 14, 18, 22, and 26, which are all unbraced vertical posts directly connected to the top chord of the bridge. In a pure truss model, these vertical members are termed zero-force members [56] and carry no load. Railroad truss bridges often function as space-frame structures, with members primarily carrying axial forces [46]. Herein, FE analysis of the Calumet 2D bridge model revealed that the forces in the vertical members can be considered to be zero, in comparison to other structural members. As a result, damage induced on these vertical members is not observable in the measured responses. This prior knowledge can be integrated into the updating pipeline using gradient scaling, as detailed in Section 4.5.

Additionally, sparsity is incorporated into the damage identification approach through L1 regularization, based on the assumption that not all members are likely to be damaged. The loss function in Equation (8) is modified to include an L1-MSE term for the learnable deviation ratios, expressed as

$$l_1^{reg} = \lambda_{l_1} \sum_{i=1}^{N_m} |k_i^d - 1.0| \qquad (9)$$

where, $\lambda_{l_1}$ is the L1 regularization constant, and $N_m$ is the number of structural members considered for updating. Thus, the total loss for the optimizer is obtained as

$$l_n^{total} = l_n^{combine} + l_1^{reg} \qquad (10)$$

For the case study presented, a scaling factor of 1e4 was selected based on trial-and-error, with $\lambda_{l_1}$ set to 1e-3 times the average error in the primary objective function (Equation (8)). Further tuning demonstrated that the RMSprop optimizer outperformed AdamW in reducing false positives and improving the overall accuracy. Specifically, using RMSprop with a lower learning rate of 3e-3, along with constraints based on prior information, the PINN based model predicted deviation ratios of [0.702, 1.244, 0.806], achieving an average accuracy of 99.50%, as shown in Figure 5.4(a). Figure 5.4(b) highlights that there were zero false positives, with a maximum error of 0.8%. However, convergence required 300 epochs with RMSprop, compared to 200 epochs when using AdamW. Herein, when using both the acceleration and displacement response at free DOFs for the loss function in Equation (8), the model achieves faster convergence, within 150 epochs. Notably, the damage identification and updating process can operate solely using the error metric based on displacement data. The results in the following section are obtained using only displacement response for the loss, demonstrating the model's efficacy in working with limited measurements on the bridge. The model performance under the presence of noise in simulated responses is validated in the next section.

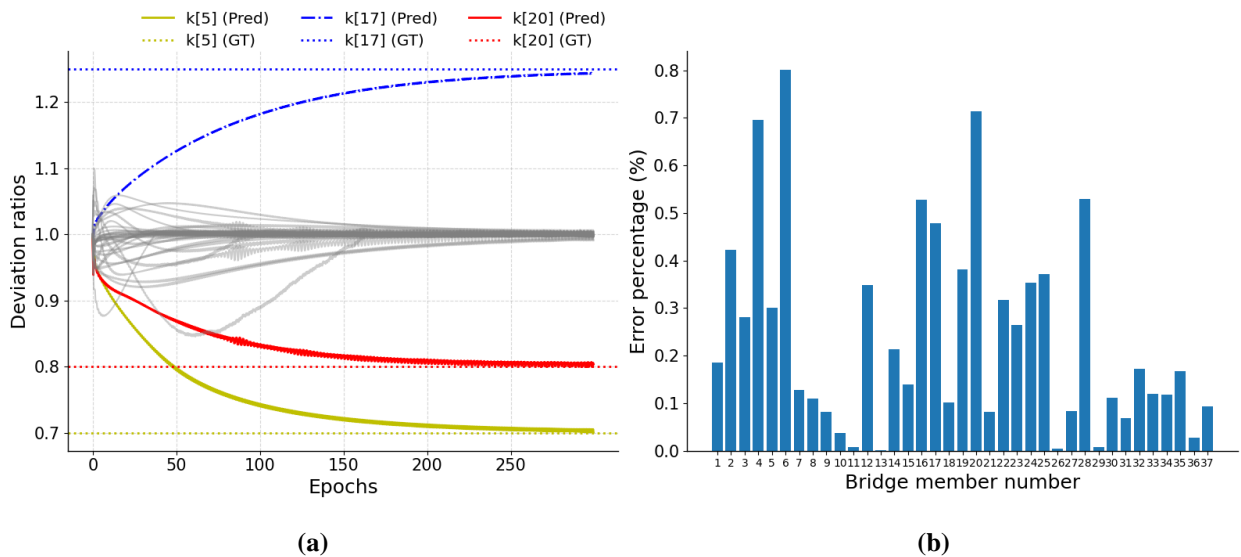

(a) (b)

**Figure 5.4 PINN-based model with RMSprop and prior: (a) evolution of deviation ratios during training, (b) bar plot of percentage error in deviation ratios for all the members.**

*5.2.3 Performance with noise in the data*

Sensor measurements in real-world scenarios inevitably contain noise, which poses challenges for accurate damage identification. For the three-member damage case analyzed in Section 5.4.1, a 5% Gaussian noise was introduced into the simulated displacement responses for the Calumet bridge 2D model, as shown in Figure 5.5(a), representing typical noise levels encountered in practice. Despite the noisy data, the training progress plot in Figure 5.5(b) shows that the PINN based model successfully converged, identifying the three damaged members ([0.709, 1.242, 0.804]) with zero false positives and a maximum error of 1.27%. The next subsection presents the results of damage identification across a range of damage scenarios.

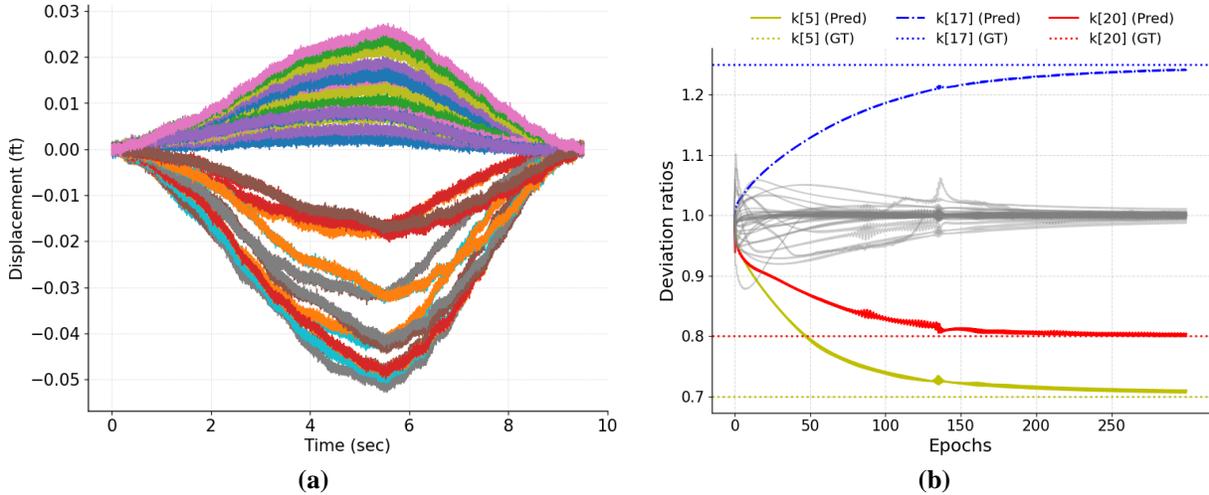

**Figure 5.5:** Damage identification with noise: (a) displacement time history with 5% noise for train crossing at 50 MPH, (b) training progress plot.

*5.2.4 Quantitative evaluation across multiple damage cases*

The performance of the PINN based model is further evaluated across various damage cases on the 2D Calumet bridge model. These cases include scenarios with varying numbers of damaged members and different locations of damaged members selected on the bridge model. Additionally, deviation ratios for the damaged members are randomly sampled to introduce variability. Table 1 summarizes the results obtained for ten damage scenarios, with cases also chosen to include damage in critical members, based on the FE model, and their structural behavior (tension or compression). Deviation ratios are randomly sampled from a normal distribution within the range of 0.6 to 1.6. The evaluations are conducted over 300 epochs using the RMSprop optimizer with a learning rate of 3e-3, employing SmoothL1Loss on the displacement responses as the loss function. The results demonstrate high accuracy across all cases, with a minimum accuracy of 98.75% and zero false positives. The maximum error observed is 2.23%. These findings confirm the model's robustness and accuracy across different damage scenarios. The next section discusses the results of the proposed strategy applied to the 3D Calumet bridge model.

Table 1. Damage simulations for 2D Calumet bridge model

| Case | Damaged members | Ground truth (GT) deviation ratios / Predicted (Pred) deviation ratios | Average accuracy (%) | False-positive / Max error (%) |
|---|---|---|---|---|
| 1 | 5, 17, 20 | [0.7, 1.25, 0.8] / [0.702, 1.244, 0.806] | 99.50 | None / 0.80 |
| 2 | 11, 19, 27 | [0.85, 1.15, 0.7] / [0.851, 1.151, 0.702] | 99.84 | None / 0.52 |
| 3 | 3, 11, 20, 24, 31 | [0.8, 1.25, 0.7, 0.9, 0.75] / [0.806, 1.249, 0.702, 0.905, 0.752] | 99.62 | None / 0.72 |
| 4 | 1, 10, 32, 35, 3, 20 | [0.9, 0.65, 0.85, 1.15, 0.7, 1.3] / [0.903, 0.650, 0.848, 1.137, 0.699, 1.301] | 99.70 | None / 1.22 |
| 5 | 8, 15, 17, 24 | [1.11, 1.29, 1.16, 1.52] / [1.106, 1.294, 1.158, 1.527] | 99.66 | None / 0.52 |
| 6 | 9, 30 | [1.27, 0.76] / [1.243, 0.762] | 98.75 | None / 2.16 |
| 7 | 1, 11, 10, 29, 3, 8, 20 | [1.11, 1.17, 0.94, 0.92, 1.18, 1.08, 1.2] / [1.106, 1.169, 0.941, 0.921, 1.174, 1.079, 1.199] | 99.82 | None / 0.67 |
| 8 | 11, 16, 31, 20, 33, 34, 23, 17, 37, 29 | [0.76, 1.16, 1.19, 0.63, 0.85, 0.88, 0.87, 0.93, 0.89, 1.08] / [0.761, 1.164, 1.181, 0.634, 0.852, 0.882, 0.872, 0.929, 0.892, 1.080] | 99.67 | None / 2.23 |
| 9 | 12, 16, 20, 24, 28 | [0.92, 1.09, 0.86, 1.13, 1.17] / [0.921, 1.088, 0.862, 1.133, 1.167] | 99.80 | None / 0.48 |
| 10 | 13 | [0.75] / [0.751] | 99.91 | None / 0.53 |

## 5.3. Results on 3D Calumet bridge model

This section validates the performance of the PINN-based model on the 3D Calumet bridge model. Herein, the damage identification study utilizes the Guyan reduced model of the 3D Calumet bridge and the implicit RK formulation within the Phy-RNN cell, as described in Sections 4.4 - 4.5. Damage in railroad truss bridges often occurs in clusters, particularly in regions experiencing high stress concentrations, fatigue damage, and larger transverse displacements, such as the central region of the bridge. The AREMA's Bridge Inspection Handbook [57] highlights that bridge inspectors prioritize these regions during inspections. Two damage clusters are considered in this study: (a) Cluster A – central region: due to larger vibrations near the central region, damage often propagates from the cross-bracing at rail level, which have smaller cross-sections, towards the connecting main structural members. Herein, this cluster includes four damaged members: a cross-bracing, a vertical structural member, and two other diagonal structural members, as shown in Figure 5.5(a). (b) Cluster B – approach region: structural members near the approach of railroad bridges are prone to clustered damage due to external factors, including oil spills, uneven track-to-bridge transitions, and sudden train wheel impacts [58]. The damage case considered for the approach cluster here for the 3D bridge model is depicted in Figure 5.5(b).

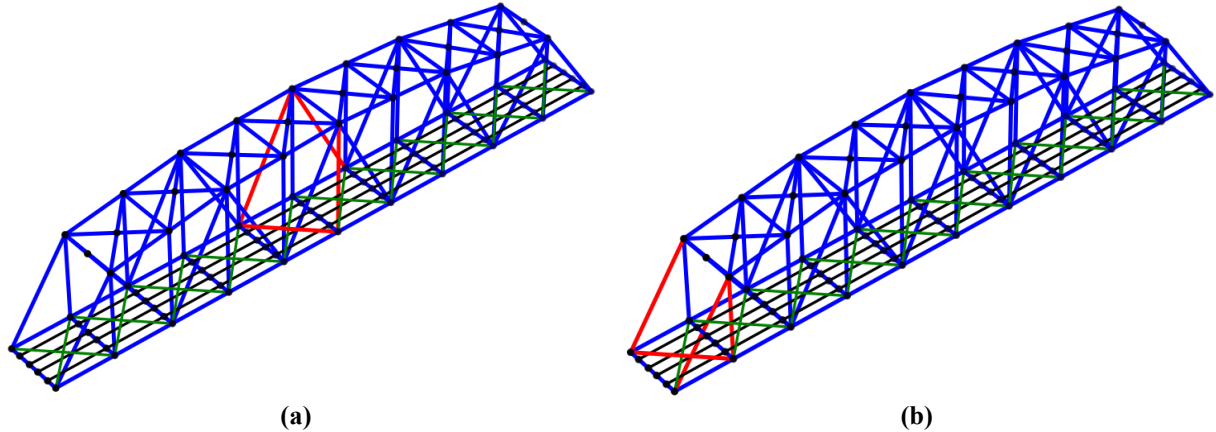

**Figure 5.5: Damage clusters for 3D Calumet bridge model: (a) Cluster A – central region, (b) Cluster B – approach region**

The PINN based model herein was initialized with the RMSprop optimizer and a CyclicLR scheduler, cycling learning rates between 5e-4 to 1e-3 over a step size of 50 epochs; a batch size of 64 was selected. A timestep of 0.002 secs was used for the implicit RK integrator within the Phy-RNN cell. The model uses the displacement responses at the selected master DOFs of the Guyan reduced model (Section 4.4) of the 3D Calumet bridge. Preliminary trials revealed that vertical members, cross-bracing, and lateral bracing exhibited low sensitivity to updates and were frequently identified as false positives. This behavior aligns with the cues from the FE model, where some of the vertical members act similar to zero-force members in a truss. Furthermore, bracing, due to their significantly smaller cross-sectional areas, shows reduced sensitivity to structural response due to damage in the members. Damage identification for less sensitive members poses a challenge. To address this, prior knowledge of the bridge model is incorporated into the PINN based pipeline using a gradient scaling factor of 1e3 for the less sensitive structural members. This study simulates a freight train crossing the bridge at 80 MPH, resulting in a displacement response time history duration of 28.87 seconds. The global stiffness matrix of the Guyan reduced model of the 3D Calumet bridge model has a size of 136x136, and the displacement response matrix for the observed DOFs has a size 14400 x 126, visualized in Figure 5.6(a).

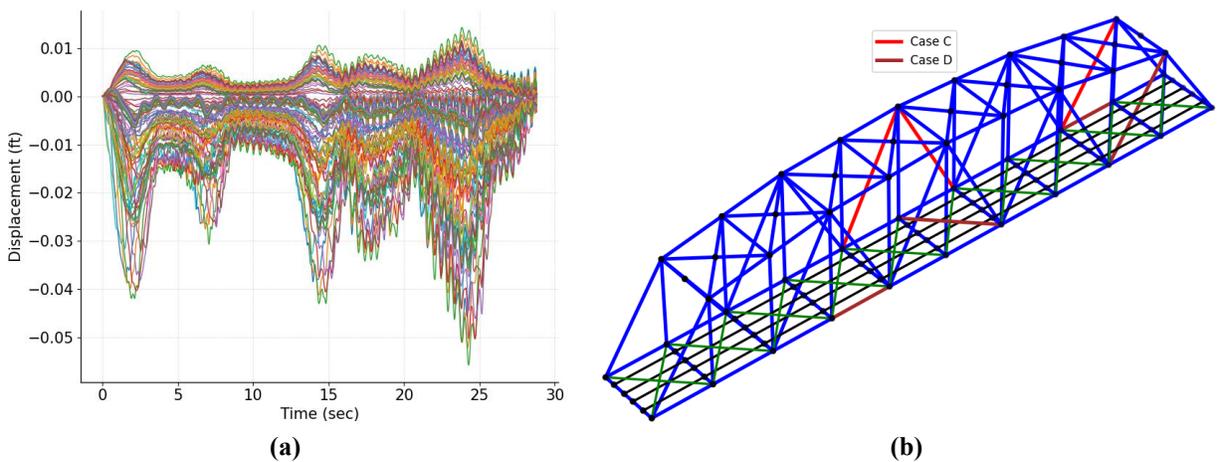

**Figure 5.6: (a) Simulated displacement time history for train crossing at 80 MPH for Cluster A, (b) Damaged members for Case C and Case D**

The deviation ratios and the corresponding damaged members for Cluster A and Cluster B are summarized in Table 2. In both cases, the model accurately identifies all damaged members, showing strong correlation between ground-truth (GT) and predicted deviation ratios, with an average accuracy exceeding 98%. For Cluster A, the training progress plot in Figure 5.7(a) illustrates the evolution of deviation ratios during training for the 157 structural members on the 3D Calumet bridge model. In Figure 5.7(a), gray-colored lines represent healthy members, while the other colored lines indicate the damaged members. The PINN based model required approximately 400 epochs to achieve convergence for the 3D Calumet bridge. For this damage cluster, the maximum error in the deviation ratios of the updated 3D bridge model is 7.16%, with two false positives, as seen from the error plot in Figure 5.7(b). In Cluster B, there are two false positives with an error of 6.31% and 7.10%. To further demonstrate the model's effectiveness in damage identification, two additional damage scenarios with randomly selected structural members on the 3D Calumet bridge are simulated, illustrated in Figure 5.6(b), and are described in Table 2. Overall, Table 2 illustrates that the proposed PINN based model performs well on identification of damage clusters which includes less-sensitive members of the bridge, and randomly selected structural members at different regions of the Calumet bridge.

Table 2. Damage simulations for Calumet 3D bridge model

| Damage scenario | Damaged members | Ground Truth (GT) deviation ratios / Predicted (Pred) deviation ratios | Average accuracy (%) | False-positive members / Max error (%) |
|---|---|---|---|---|
| Cluster A | 19, 21, 57, 127 | [0.75, 0.86, 0.83, 0.91] / [0.751, 0.865, 0.832, 0.929] | 99.23 | [126, 130] / 7.16, 5.57 |
| Cluster B | 65, 66, 29, 134 | [0.7, 0.86, 0.9, 0.8] / [0.705, 0.872, 0.900, 0.848] | 98.06 | [126, 130] / 6.31, 7.10 |
| Case C | 13, 19, 21 | [0.7, 1.25, 0.8] / [0.702, 1.25, 0.804] | 99.74 | [26] / 5.29 |
| Case D | 2, 50, 125, 44 | [0.88, 0.78, 0.82, 0.91] / [0.875, 0.782, 0.861, 0.921] | 98.32 | [126, 130] / 7.89, 6.05 |

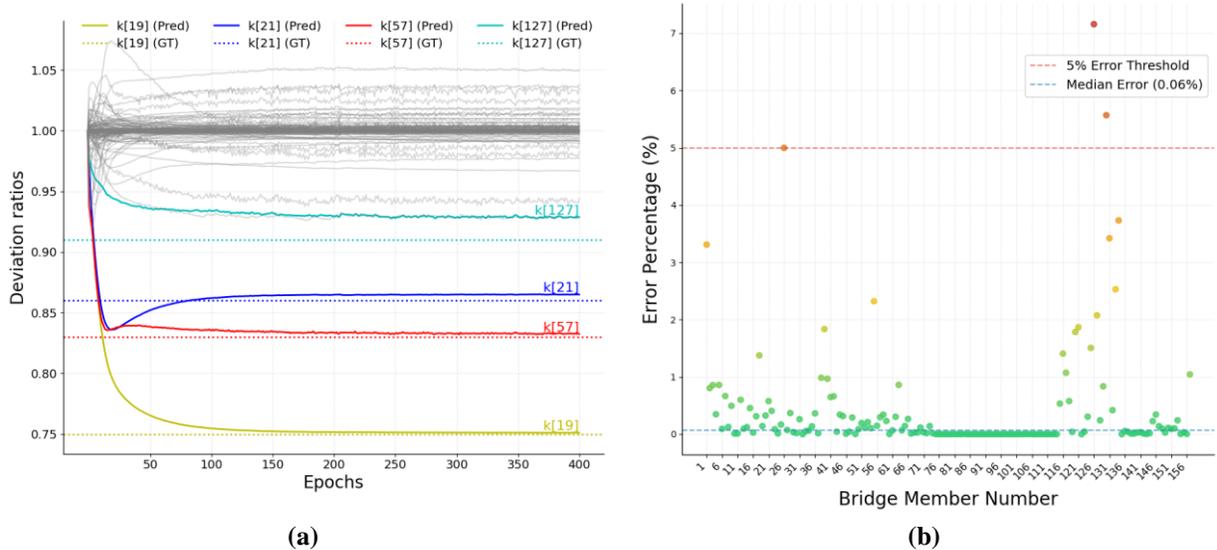

(a) (b)

Figure 5.7: Damage Cluster A: (a) Training progress plot, and (b) deviation ratio error plot

## 6. DISCUSSION

While the proposed PINN based model shows promising results, there are several challenges that remain open for future research. In this study, the false positives are identified based on a deterministic error threshold of 5% from the healthy state. Probabilistic damage identification with a confidence score associated with damage parameters would be explored in a future study. Herein, damage identification for the 3D Calumet bridge model is performed with the Guyan reduced model. The selection of DOFs influences damage identification results, ideally, the DOFs to be selected for measuring responses need to be optimized such the number of sensors or DOFs monitored on the bridge can be minimized for cost-savings and efficiency, while capturing the overall behavior of bridge for effective damage detection. This study focuses on the feasibility of unsupervised PINNs for simultaneous damage identification and model updating in LTV railroad bridge systems, rather than benchmarking against existing methods. A main computational bottleneck for the PINN based model lies in the RK integrator. Currently, work is ongoing to use semi-supervised PINN architecture to substitute for the differentiable RK integrator to make the deep learning pipeline more efficient. Additionally, this paper focuses on modeling damage in structural members as a reduction in stiffness. However, other types of damage, such as support settlement and damage to connections or joints, are also critical for effective railroad bridge management and will be addressed in future studies. This study uses simulated scenarios of damage on a full-scale railroad truss bridge for validation of the PINN based damage identification approach. Future research will address validation through experimental tests on an instrumented test structure. Finally, ongoing work on automated, vision-based damage localization and severity prediction from drone surveys on steel truss bridges will be integrated into the proposed PINN based damage identification and model updating pipeline.

Moving forward towards the goal of creating robust digital twins for monitoring critical bridge assets, the approach proposed in this study offers a promising pathway. By utilizing data from sensors deployed on a railroad bridge for long-term monitoring, the method allows the bridge model to remain updated based on observed data. Notably, the proposed PINN based model requires response measurements from only a single train crossing event to update the bridge model and perform damage identification. This capability potentially makes the approach suitable for campaign-style monitoring deployments, enabling efficient railroad bridge assessment during routine inspections by bridge inspectors.

## 7. CONCLUSION

A physics-informed neural network (PINN) based approach was developed for damage identification in railroad truss bridges. The approach incorporated the governing differential equations of the linear time-varying bridge-train system dynamics directly into a deep learning model, using a recurrent neural network (RNN) based architecture with a custom Runge-Kutta (RK) integrator cell. An unsupervised learning approach utilized train wheel load data and bridge responses from train crossings to update the bridge finite element model effectively. This method also quantified damage severity and localized affected structural members without requiring extensive labeled datasets.

Case studies on the simulated railroad truss Calumet Bridge in Chicago, IL demonstrated the robustness of the proposed strategy in identifying damage, with low false-positive rates even with noisy data. For the simplified 2D Calumet bridge model, the approach achieved an average accuracy exceeding 98% across multiple damage scenarios, with randomly selected structural members and varying damage levels. Further validation on the 3D Calumet bridge model was conducted with a Guyan reduced formulation to manage large degrees of freedom efficiently, achieving over 98% accuracy in damage identification and less than 2% false positives across scenarios, including damage clusters representative of real-world conditions. This performance underscored the approach's potential for accurate, context-aware model updating, particularly when supplemented with available site inspection and prior information from FE models. The results highlight the potential of the proposed PINN based approach in advancing structural health monitoring and maintenance practices for railroad bridges.

## 8. ACKNOWLEDGEMENTS

The authors would like to thank Professor Fernando Moreu for his valuable suggestions and guidance. They also extend their gratitude to Professor Robin Eunju Kim for sharing the data related to the Calumet Bridge.

## 9. DECLARATION OF CONFLICTING INTERESTS

The author(s) declared no potential conflicts of interest with respect to the research, authorship, and/or publication of this article.

## 10. FUNDING

The author(s) disclosed receipt of the following financial support for the research, authorship, and/or publication of this article: The data collection for this research was funded by the Federal Railroad Administration under the BAA 2010-1 project entitled 'Campaign Monitoring of Railroad Bridges in High-Speed Rail Shared Corridors using Wireless Smart Sensors,' Contract No. DTFR53-13-C-00047.

## 11. DATA AVAILABILITY

The data supporting this study would be made available from the authors upon reasonable request.